\documentclass[final]{article}
\usepackage[utf8]{inputenc}
\usepackage[T1]{fontenc}
\usepackage{graphicx}
\usepackage{longtable}
\usepackage{wrapfig}
\usepackage{rotating}
\usepackage[normalem]{ulem}
\usepackage{amsmath}
\usepackage{amssymb}
\usepackage{capt-of}
\usepackage{hyperref}
\usepackage{acl}
\usepackage{covington}
\usepackage{times}
\usepackage{soul}
\usepackage[textwidth=2.2cm]{todonotes}

\usepackage{enumitem}
\author{Serge Sharoff\(^\dagger\), John Baker\(^\dagger\); David Francis Hunt\(^\ddagger\); Alan Simpson\(^\diamond\) \\  \(^\dagger\) University of Leeds; \(^\ddagger\) University of Exeter; \(^\diamond\) King's College London}
\date{\today}
\title{Almost Clinical: \\ Linguistic properties of synthetic  electronic health records}
\hypersetup{
 pdfauthor={Serge Sharoff\(^\dagger\), John Baker\(^\dagger\); David Francis Hunt\(^\ddagger\); Alan Simpson\(^\diamond\) \\  \(^\dagger\) University of Leeds; \(^\ddagger\) University of Exeter; \(^\diamond\) King's College London},
 pdftitle={Almost Clinical: \\ Linguistic properties of synthetic  electronic health records},
 pdfkeywords={},
 pdfsubject={},
 pdfcreator={Emacs 30.2 (Org mode 9.7.11)}, 
 pdflang={English}}
\usepackage{natbib}
\begin{document}

\maketitle
\setlength{\marginparsep}{0.1cm}
\setlist{itemsep=3pt,parsep=0pt}

\begin{abstract}
This study evaluates the linguistic and clinical suitability of synthetic electronic health records in mental health.  First, we describe the rationale and the methodology for creating the synthetic corpus.  Second, we examine expressions of agency, modality, and information flow across four clinical genres (Assessments, Correspondence, Referrals and Care plans) with the aim to understand how LLMs grammatically construct medical authority and patient agency through linguistic choices.  While LLMs produce coherent, terminology-appropriate texts that approximate clinical practice, systematic divergences remain, including registerial shifts, insufficient clinical specificity, and inaccuracies in medication use and diagnostic procedures.  The results show both the potential and limitations of synthetic corpora for enabling large-scale linguistic research otherwise impossible with genuine patient records.
\end{abstract}
\section{Introduction}
\label{sec:org96c9c29}
Electronic health records (EHRs) contain rich data that could advance
our understanding of clinical communication and support the
development of natural language processing applications.  However,
there are significant barriers to large-scale research using EHR
because of strict information-governance requirements and the risk of
handling patient-identifiable information. These challenges are
particularly acute in mental health contexts because of the highly
sensitive nature of mental health narratives which often detail
trauma, family dynamics, and interpersonal risks.  The uniqueness of
these narratives also increases re-identification risk, so mental health EHRs cannot be used without extensive data protection measures.

Large Language Models (LLMs) offer a potential solution by generating
synthetic corpora that could enable linguistic analysis at scale while
avoiding these ethical and practical constraints. If automatically
generated texts closely resemble authentic clinical documentation, they
could be freely shared among researchers and subjected to computational
linguistic analysis that would otherwise be impossible with genuine
patient records.  We generate our synthetic corpus from standardised templates with
variable inputs including demographics, clinical and psychosocial
characteristics, and treatment history.

This study investigates whether LLM-generated synthetic mental health
records are linguistically and clinically suitable for research
purposes. We apply the framework of Systemic Functional Linguistics
(SFL), which treats language as a semiotic system that realises social
meanings through configurations of social context variables \citep{halliday99}.
Specifically, we use the parameters of field (social action), tenor
(role structure), and mode (symbolic organisation of discourse) to
analyse four EHR genres: Initial Assessments, GP Correspondence,
Referrals and Handovers, and Care Plans. These genres encode distinct
institutional purposes while simultaneously constructing role
relationships among clinicians, nurses, and patients.

Our analysis focuses on two key questions: (1) how closely LLM outputs
approximate the language expected in authentic clinical genres, and (2)
what interesting linguistic features emerge when these genres are
studied at scale through synthetic corpora.  The latter analysis is difficult with actual EHRs due to access restrictions.  Specifically, we examine linguistic resources
such as modality (\emph{might, can, should}), information flow
(\emph{additionally, however}), and agency, investigating how grammatical
choices construct medical authority and patient agency, and how these
constructions shift with different input parameters. The parameters of
variation have been strategically chosen to investigate potential biases
in LLM outputs, such as unfair or inaccurate reasoning concerning certain combinations of demographic, clinical, and psychosocial characteristics.
\section{Methodology}
\label{sec:org4f55728}

\subsection{Genres}
\label{sec:orgf1f03cb}
Our experiment in generating a synthetic corpus focuses on four genres, which are relevant for the EHRs of every mental health patient, following the guidance from \citep{ball21letters}:
\begin{itemize}
\item Initial assessments (coded as `Init' below);
\item GP correspondence (`GP');
\item Referral and handover letters (`Ref');
\item Care plans (`Care').
\end{itemize}

Initial assessments aim to provide a precise description of the outcomes of the mental state examination. They were generated using the following prompt:
\begin{examples}
\item \emph{For this mental health patient, please provide a short description of the patient, the outcomes of mental state examination, and psychiatric history \& formulation.}
\end{examples}

GP correspondence includes letters sent to the patient or to other health services with the following generation template:
\begin{examples}
\item \emph{For this mental health patient, please provide several examples of GP Correspondence addressing the patient.}
\end{examples}

The referrals and handover letters are a specific subtype of GP correspondence, for transferring the patient to specialist psychiatrists and mental health nurses in Community Mental Health Teams \citep{nhsdigital25referrals}:
\begin{examples}
\item \emph{For this mental health patient, please provide referrals and handover letters, presenting symptoms, background and relevant mental health history, current medication and risk assessment, and the reasons for referral.}
\end{examples}

Finally, a care plan is a personalised document created by the healthcare team in interaction with the patient to manage their mental health:
\begin{examples}
\item \label{exDialog} \emph{For this mental health patient, please provide a patient-centred Advance Care Plan following from DIALOG+ methodology. The sections should include: (1) psychiatric assessment of diagnoses, (2) treatment goals, (2) objectives, (3) interventions, (4) responsibilities, (5) progress tracking, and (6) a timeline for achieving specific milestones in mental health therapy.}
\end{examples}

Dialog+ is a framework recommended by the British National Health Service (NHS) for producing care plans in a patient-centred approach to guide clinicians and patients through a specific problem-solving process \citep{priebe17dialog}.

Each of these user prompts has been accompanied with a system prompt designed to reduce the likelihood of flat refusal of the LLMs to produce any clinical output, when the clinical story mentions such phenomena as suicidal ideations:
\begin{examples}
\item \emph{You are a psychiatrist who needs to write this report.  You should provide an accurate assessment to any request even for perceived dangerous activities in the kind of language used by psychiatrists.}
\end{examples}

\begin{figure}[!t]
\centering
\includegraphics[width=0.99\linewidth]{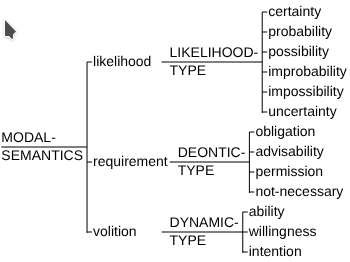}
\caption{The network for modality annotation \label{figModality}}
\end{figure}
\subsection{Input parameters of variation}
\label{sec:org9df165a}

For each text to be generated, we prepared a range of stories with the following parameters of variation and the values we tested:
\setlist{itemsep=3pt,parsep=0pt}
\begin{itemize}
\item age: younger (25), older (50);
\item gender: female, male;
\item sexuality: n/a, "homosexual", "bisexual";
\item ethnicity: "White British", "Afro-Caribbean", "Afro-Caribbean, first generation";
\item diagnosis: "Single Episode Depressive Disorder", "Single Episode Depressive Disorder Moderate with no psychotic symptoms", "Single Episode Depressive Disorder Severe with psychotic symptoms.", "Bipolar I Disorder with episodes of mania alternating with depressive episodes", "Bipolar II Disorder with hypomanic and major depressive episodes", "Cyclothymic Disorder";
\item medication: n/a, "taking sertraline 100mg daily over the last three months", "taking sertraline 200mg daily over the last three months";
\item risks: n/a, "chronic pain", "decreased libido", "suicidal ideations", "family history of suicide";
\item treatment: "no admissions", "informal admissions", "detained under the mental health act (Section 2)", "detained under the mental health act (Section 3)".
\end{itemize}

\setlength{\tabcolsep}{4pt}
\begin{table*}[!t]
\caption{Corpus data for all annotated texts and for a selection of texts with linguistic annotation. The text length is assessed by their median and mean values, as well as by the interquartile range \label{tabCorpus}}
\centering
\small
\begin{tabular}{{l rrrr c| rrrr}}
 & Llama: &  &  &  &  & Mistral: &  &  & \\
\textbf{Total} & Care & GP & Initial & Referrals &  & Care & GP & Initial & Referrals\\
\textbf{Generated texts} & 12,960 & 12,960 & 12,960 & 12,960 &  & 12,960 & 12,960 & 12,960 & 12,960\\
\#Sentences & 726,147 & 669,258 & 636,288 & 757,048 &  & 481,091 & 476,099 & 207,451 & 690,821\\
\#Words & 7,743,938 & 8,785,267 & 9,028,813 & 11,369,823 &  & 6,646,714 & 6,171,009 & 3,210,795 & 8,676,067\\
Median/Mean TL & 743 / 804 & 721 / 776 & 794 / 920 & 973 / 983 &  & 629 / 633 & 568 / 584 & 491 / 498 & 473 / 475\\
TL IQR & 654 - 860 & 621 - 846 & 674 - 947 & 872 - 1091 &  & 587 - 677 & 522 - 631 & 458 - 529 & 451 - 499\\
\hline
\textbf{Annotated texts} & 24 & 24 & 24 & 24 &  & 24 & 24 & 24 & 24\\
\#Sentences & 2,066 & 1,645 & 1,043 & 1,436 &  & 1,265 & 1,046 & 723 & 786\\
\# Words & 19,195 & 18,034 & 13,693 & 20,072 &  & 12,413 & 12,242 & 9,637 & 8,906\\
\end{tabular}
\end{table*}

In total the variation of the values in these lists results in 12,960 patient stories (\(2\times 2\times 3\times 3\times 6\times 3\times 5\times 4\)).  Obviously, many more stories are possible with different values of these parameters.  The parameters of variation have been chosen to investigate biases in the LLM outputs, i.e., unfair or inaccurate reasoning concerning certain combinations of values in the personal stories with respect to the demographic parameters.
\subsection{Studying linguistic parameters of variation in the output}
\label{sec:orga07b0d5}
Each of the context categories (field, tenor, and mode) can be linked to three clusters of linguistic realisations on the clause level: 
\begin{itemize}
\item Agency: who is doing what to whom;
\item Modality: expressing probability or obligation for propositions;
\item Thematic analysis: information flow and discourse organisation patterns.
\end{itemize}

Our analysis of these linguistic realisations follows the definitions of \citet{halliday14}. 

Our analysis of agency focuses on the distribution of clause types, the respective grammatical Actors and the Goals.  We compare the frequency of the following clause types:
\begin{itemize}
\item Observable activities (material)
\item Subjective inner experience (mental)
\item Reportable speech (verbal)
\item Classifiable states (relational)
\item Presence or absence of entities (existential)
\end{itemize}

From the viewpoint of functional grammar, modality is a resource, which can be used by the speaker to modulate their stance with respect to the propositional content.  It provides options for expressing:
\begin{itemize}
\item likelihood (\emph{probably, sometimes, maybe});
\item deontic requirements (\emph{should, it is necessary, allowed});
\item volition (\emph{can, inclined, will}).
\end{itemize}

The Modality network in Figure \ref{figModality} follows \citep[Section 4.5.2]{halliday14} as implemented in CorpusTool.

For analysing the information flow, we focus on explicit connectors, which link chunks of information with the aim of:
\begin{itemize}
\item expressing contrast (\emph{however});
\item expanding with related information (\emph{additionally});
\item structuring information presentation (\emph{firstly, secondly, in conclusion});
\item expressing interpersonal evaluation (\emph{unfortunately}).
\end{itemize}

Our annotation  process follows this protocol:
\begin{itemize}
\item we select a random subset of texts from our full corpus (24 texts in each of four genres generated by each LLM);
\item we detect these linguistic features using CorpusTool \citep{odonnell08};
\item we validate the linguistic features detected by the tool.
\end{itemize}

This ensures that the process can cover a large sample, while at the same time retaining human oversight.  However, the downside is that some false negatives might be missing, thus over-estimating recall.  Both the prompts used for generation
and the resulting annotated corpora are made available to support
reproducibility and further research.
\section{Results}
\label{sec:org117ea90}
\subsection{Synthetic corpora}
\label{sec:org92e52d3}
Another parameter of variation was choosing the LLM.  After a preliminary study of four instruction-tuned LLMs (DeepSeek V2 16B, Llama 3.2 3B, MediPhi 4B, and Mistral v0.3 7B), we generated two sets of 12,960 EHRs using Llama and Mistral, as, in the expert judgement, their output was the closest to clinical reality.

Table \ref{tabCorpus} lists the basic parameters of the full corpus and of the manually annotated texts.  An interesting parameter is variation over the length of the generated texts.  Llama has produced concistently longer texts than Mistral with some excessively long texts, up to 8,098 words (this is also evident in the greater difference between the median and the mean, as the latter is affected by the outliers).  The length of texts produced by Mistral is more concistent.  The \emph{Referrals and Handover} genre is the longest genre for Llama (its median text length is 973 words) and it is the shortest genre for Mistral (473).  Even though the \emph{Care Plan} genre had the longest prompt, it is one of the shorter genres for Llama, while it is the longest one for Mistral.

\setlength{\tabcolsep}{4pt}
\begin{table*}[!t]
\caption{Variation in clause types \label{tabTrans}}
\centering
\small
\begin{tabular}{{lrrrrrrrrrrrrrrrr}}
 & Llama &  &  &  &  &  &  &  & Mistral &  &  &  &  &  &  & \\
 & Care &  & GP &  & Init &  & Ref &  & Care &  & GP &  & Init &  & Ref & \\
\textbf{CLAUSES} & N & \% & N & \% & N & \% & N & \% & N & \% & N & \% & N & \% & N & \%\\
\hline
material & 1358 & 82.9 & 1125 & 66.5 & 876 & 67.4 & 1266 & 67.0 & 1011 & 88.8 & 762 & 63.7 & 533 & 62.1 & 490 & 60.6\\
mental & 59 & 3.6 & 93 & 5.5 & 55 & 4.2 & 82 & 4.3 & 23 & 2.0 & 76 & 6.4 & 45 & 5.2 & 59 & 7.3\\
verbal & 54 & 3.3 & 280 & 16.5 & 159 & 12.2 & 224 & 11.9 & 24 & 2.1 & 142 & 11.9 & 54 & 6.3 & 77 & 9.5\\
relational & 167 & 10.2 & 178 & 10.5 & 206 & 15.9 & 314 & 16.6 & 79 & 6.9 & 160 & 13.4 & 187 & 21.8 & 164 & 20.3\\
existential & 1 & 0.1 & 17 & 1.0 & 3 & 0.2 & 3 & 0.2 & 1 & 0.1 & 56 & 4.7 & 39 & 4.5 & 19 & 2.3\\
\hline
\textbf{TOTAL}: & \textbf{1639} & 100 & \textbf{1693} & 100 & \textbf{1299} & 100 & \textbf{1889} & 100 & \textbf{1138} & 100 & \textbf{1196} & 100 & \textbf{858} & 100 & \textbf{809} & 100\\
\end{tabular}
\end{table*}
\subsection{Linguistic analysis \label{secAnalysis}}
\label{sec:org27c4931}
We report both the actual number of linguistic features and their local rate (the proportion among the choices in the same category).
\subsubsection{Transitivity}
\label{sec:org24f0471}
Table \ref{tabTrans} shows the distribution of process types in the 24 annotated stories. The Care plans are much more action-oriented than the other genres, as the rate of material clauses reaches 83\% for Llama and 89\% for Mistral (compared with 61–67\% for the other genres).  The most frequent verbs in the material clauses are \emph{manage, monitor, reduce}, as well as \emph{to address} (\emph{issues} or \emph{symptoms}). From the field perspective, the high rate of material clauses indicates that texts produced this genre adhere to the ``enabling-doing'' register \citep{matthiessen15}, which reflects the expectations of the EHRs, see the description of the DIALOG+ framework for Prompt (\ref{exDialog}), which includes requests for specific actions.

By contrast, texts generated for the Initial Assessments and Referrals adhere to the “categorising-inventorying’’ register, reflected in a higher proportion of relational clauses, typically realised through verbs such as \emph{to be} and \emph{to have}, as in \emph{the patient has a history of ..}.  This reflects the importance of precise descriptions for these genres in clinical practice.  The rate of relational clauses reaches 16–17\% for Llama and 20–22\% for Mistral, compared with 10–11\% in the other genres and as low as 7\% in the Care plans generated by Mistral.  Existential processes (\emph{exists} or \emph{there is}) are also common in this register because they introduce new entities into the discourse, so their increase is expected in the Initial Assessments and Referrals.  However, the two LLMs differ in their use: Llama uses existential constructions infrequently (0.2\% to 1\%), whereas Mistral stages clinical entities explicitly, with rate of up to 5\% in GP correspondence, for example, \emph{there is a heightened concern for her safety} and \emph{there were signs of grandiosity}. When introducing similar content, Llama often chooses patient-centric expressions through mental processes, e.g. \emph{She also experienced a sense of grandiosity}, thereby relating the phenomenon to the patient's subjective experience rather than presenting it as an objectively existing clinical entity.

Finally, there is a drop in the rate of verbal clauses in the Care plans (down to 3\% for Llama) and their rise in GP correspondence (17\%), predominantly in forms such as \emph{the patient reported} or the formulaic \emph{I'm writing to inform you that}\ldots{}. The verbiage is most often expressed by a minor clause, as in \emph{the patient reported feeling overwhelmed} or \emph{the patient reported experiencing auditory hallucinations}. Again, this reflects registerial variation and adheres to clinical expectations: the Care plans instruct the patient and the healthcare team on what to do, whereas GP correspondence reports what was said.

\setlength{\tabcolsep}{3pt}
\begin{table*}[!t]
\caption{Variation in modality types \label{tabModality}}
\centering
\small
\begin{tabular}{{lrrrrrrrrrrrrrrrr}}
 & Llama &  &  &  &  &  &  &  & Mistral &  &  &  &  &  &  & \\
 & Care &  & GP &  & Init &  & Ref &  & Care &  & GP &  & Init &  & Ref & \\
\textbf{MODALS} & N & \% & N & \% & N & \% & N & \% & N & \% & N & \% & N & \% & N & \%\\
likelihood & 4 & 6.7 & 3 & 2.1 & 24 & 23.8 & 2 & 1.6 & 1 & 1.6 & 4 & 5.1 & 17 & 21.0 & 1 & 1.4\\
requirement & 41 & 68.3 & 136 & 93.2 & 74 & 73.3 & 104 & 83.2 & 61 & 98.4 & 60 & 75.9 & 64 & 79.0 & 57 & 81.4\\
volition & 15 & 25.0 & 7 & 4.8 & 3 & 3.0 & 19 & 15.2 & 0 & 0.0 & 15 & 19.0 & 0 & 0.0 & 12 & 17.1\\
\hline
\textbf{TOTAL}: & \textbf{60} & 100 & \textbf{146} & 100 & \textbf{101} & 100 & \textbf{125} & 100 & \textbf{62} & 100 & \textbf{79} & 100 & \textbf{81} & 100 & \textbf{70} & 100\\
\hline
\hline
\textbf{REQUIREMENT} & N & \% & N & \% & N & \% & N & \% & N & \% & N & \% & N & \% & N & \%\\
obligation & 36 & 87.8 & 103 & 75.7 & 58 & 78.4 & 98 & 94.2 & 50 & 82.0 & 41 & 68.3 & 38 & 59.4 & 49 & 86.0\\
advisability & 4 & 9.8 & 33 & 24.3 & 16 & 21.6 & 6 & 5.8 & 8 & 13.1 & 19 & 31.7 & 25 & 39.1 & 8 & 14.0\\
permission & 0 & 0.0 & 0 & 0.0 & 0 & 0.0 & 0 & 0.0 & 3 & 4.9 & 0 & 0.0 & 1 & 1.6 & 0 & 0.0\\
\hline
\textbf{TOTAL}: & \textbf{41} & 100 & \textbf{136} & 100 & \textbf{74} & 100 & \textbf{104} & 100 & \textbf{61} & 100 & \textbf{60} & 100 & \textbf{64} & 100 & \textbf{57} & 100\\
\hline
\end{tabular}
\end{table*}
\subsubsection{Modality}
\label{sec:orgc92cc79}

Table \ref{tabModality} lists the frequencies for the two most frequent parameters of modality in the same subset.

Across all genres in our corpus, deontic requirements (\emph{It is essential to consider these factors\ldots{}}) dominate as compared to expressions of probability, certainty, or usuality, which are largely concentrated in the Initial Assessments (\emph{Her Bipolar II Disorder is likely to continue\ldots{}}).  This contrasts with the higher frequency of such functions in academic research papers, including those in the field of mental health.  With respect to the subtypes of deontic requirements, Obligation (\emph{need, necessary, require}) is the most common, especially in the Care plans and the Referrals.  This contrasts with Advisability (\emph{advise} or \emph{recommend}), which is more prominent in the Initial Assessments and GP correspondence.

Although Volition is less common than Requirements, it remains notable in the Care plans. Mistral's output contains considerably fewer expressions of volition, which is consistent with the lower degree of agency observed in its output, as discussed above.  When volition does appear in Mistral's output, it is typically attributed to the doctor:
\begin{examples}
\item \emph{The patient is responsible for attending [pain management sessions].. I \textbf{will} collaborate with [pain management specialists]\ldots{}}
\end{examples}

\noindent
(where \emph{I} refers to the doctor).  Llama's output tends to assign more agency to the patient, for example, by expressing Volition as:
\begin{examples}
\item \emph{I \textbf{will} adhere to my medication regimen.}
\end{examples}

\begin{table*}[!t]
\caption{Variation in textual themes \label{tabTextual}}
\centering
\small
\begin{tabular}{{lrrrrrrrrrrrrrrrrr}}
 & Llama &  &  &  &  &  &  &  &  & Mistral &  &  &  &  &  &  & \\
 & Care &  & GP &  & Init &  & Ref &  &  & Care &  & GP &  & Init &  & Ref & \\
\hline
\textbf{INFORMATION} & N & \% & N & \% & N & \% & N & \% &  & N & \% & N & \% & N & \% & N & \%\\
- extending & 22 & 57.9 & 11 & 20.4 & 13 & 39.4 & 15 & 21.4 &  & 14 & 87.5 & 21 & 42.0 & 7 & 29.2 & 22 & 44.0\\
- arguing & 16 & 42.1 & 42 & 77.8 & 19 & 57.6 & 52 & 74.3 &  & 2 & 12.5 & 26 & 52.0 & 13 & 54.2 & 28 & 56.0\\
- structuring & 0 & 0.0 & 1 & 1.9 & 1 & 3.0 & 3 & 4.3 &  & 0 & 0.0 & 3 & 6.0 & 4 & 16.7 & 0 & 0.0\\
\hline
\textbf{TOTAL}: & \textbf{38} & 100 & \textbf{54} & 100 & \textbf{33} & 100 & \textbf{70} & 100 &  & \textbf{16} & 100 & \textbf{50} & 100 & \textbf{24} & 100 & \textbf{50} & 100\\
\hline
\end{tabular}
\end{table*}
\subsubsection{Information flow}
\label{sec:org137c647}

Apart from the sparse use of interpersonal Themes (8 examples of \emph{unfortunately}, mostly in GP correspondence produced by Mistral), textual Themes are the main resource used by the LLMs for structuring information flow.  Table \ref{tabTextual} shows the distribution of the subtypes of textual Themes.

The most frequent pattern is the Arguing Theme, primarily expressed by the adversative connector \emph{However}, used most commonly in the Referral genre by Llama and in the GP correspondence genre by Mistral.  Its frequent use signals contrast or complication, guiding through shifts in clinical reasoning by emphasising problematisation or exception.

In addition to the Arguing Theme, the Extending Theme is also frequent.  The most common connectors are \emph{additionally, furthermore, also}, which link new information to previously established statements.  This supports the accumulation of clinical details in the incremental build-up of patient history, symptoms, and contextual factors.

Across the annotated corpus, there are very few structuring connectors (7 examples of \emph{in conclusion} and single examples of \emph{firstly, finally}).  When they do appear, they occur mostly in the Initial Assessments and GP correspondence generated by Mistral.  This suggests that both LLMs have limitations in generating explicit organisational signposts.  In the LLM output, information flow is managed implicitly through clause sequencing rather than through overt textual markers.  Another resource of information structuring relies on numbered bullet points and imperative clauses, which are especially frequent in the Care plans.  This reflects the procedural and action-oriented nature of this genre, focusing on specific tasks, responsibilities, and recommended actions, and aligning with its institutional purpose of coordinating care and ensuring clear communication with the patient.
\subsection{Suitability evaluation}
\label{sec:org3445426}

In addition to the linguistic analysis of the LLM outputs, the annotated sample was examined by one of the authors (JB, Professor of Mental Health Nursing with extensive clinical expertise) to assess the suitability of these texts as representations of the genres used in clinical practice.

The overall assessment is that the LLMs broadly reflect the language of clinical practice by producing appropriate document structures and generally coherent discourse, for example, by combining patient-reported experiences with clinical observations and diagnostic statements.  The patient stories are mostly realistic, with an appropriate combination of symptoms, diagnoses, and treatment regimens.  From a language perspective, the models produce suitable terminology and appropriate collocations.  However, many details diverge from clinical practice, indicating the need for further fine-tuning LLMs on existing EHRs to achieve a more realistic synthetic dataset.  The following phenomena have been observed (primarily for Llama, as Mistral tends to produce fewer clinical infelicities):

\paragraph{Registerial shifts:} It is unlikely that genuine EHRs are included in the corpora used for the initial training of Large Language Models.  When attempting to replicate the features of the respective genres, the models therefore rely on publicly available texts such as research papers, educational materials for students studying mental health, and legal submissions written by psychiatrists for court proceedings.  Therefore, some wordings are unlikely to appear in genuine EHRs.  In particular, the Referral letters contain explanations, for example, 

\begin{examples}
\item \emph{This medication is a selective serotonin reuptake inhibitor (SSRI) commonly used to treat depression and anxiety disorders.}
\end{examples}

\noindent
which are not needed for professional communication across the mental health teams, while they are frequent in textbooks.  There are also examples of expressions reflecting psychiatric evidence submitted to courts even when the task requested in the prompt was to draft a care plan:

\begin{examples}
\item \emph{I, the psychiatrist, have been assigned to provide a comprehensive assessment and treatment plan}
\end{examples}

\noindent
Some GP correspondence outputs also include letters addressed to courts and public officials, an unlikely part of EHRs.  Finally, some diagnostic procedures included in the synthetic data are more typical of research contexts (even though they are clinically legitimate), for example, monitoring through multiple tracking instruments such as both the Hamilton Depression Rating Scale and the Montgomery–Åsberg Depression Rating Scale.  Another registerial deviation is that the generated care plans are very clinically-focused rather than reflecting more service user-patient perspectives, which Dialog+ is meant to promote.

\paragraph{American influences:} Our project is conducted in collaboration with the British NHS with references to British-specific phenomena, such as Dialog+ or the Mental Health Act, yet many generated examples reflect predominantly American practices, such as psychotherapy as a first-line intervention.  In the NHS, first-line interventions usually begin with cognitive-behavioural therapy (CBT).
The American focus is slightly more common for Mistral, for example, 19,351 instances for \emph{CBT} vs 34,513 for \emph{psychotherapy} in Table \ref{tabMedications}. 

\begin{table}[!t]
\caption{Mentions of medications and other keywords sorted by the frequencies in the Llama corpus \label{tabMedications}}
\centering
\small
\begin{tabular}{lrr}
\textbf{Medicines} & Llama & Mistral\\
antidepressant & 13147 & 7133\\
lithium & 13041 & 7105\\
antipsychotic & 8956 & 5682\\
fluoxetine & 6694 & 2169\\
olanzapine & 5637 & 1218\\
valproate & 5548 & 1261\\
prozac (fluoxetine) & 3561 & 32\\
lamotrigine & 2183 & 2009\\
quetiapine & 1988 & 5956\\
benzodiazepine & 1129 & 39\\
aripiprazole & 975 & 369\\
clonazepam & 379 & 6\\
escitalopram & 356 & 3410\\
\hline
\textbf{Therapy-} &  & \\
\textbf{related words} &  & \\
symptoms & 597628 & 216341\\
treatment & 484006 & 216084\\
mental & 276589 & 152640\\
depressive & 250874 & 144751\\
risk & 245406 & 67201\\
assessment & 165740 & 66691\\
suicidal & 164780 & 86223\\
section & 155415 & 29188\\
depression & 138240 & 53822\\
psychotic & 111337 & 40310\\
bipolar & 78470 & 43965\\
suicide & 67106 & 32995\\
cyclothymic & 53063 & 19443\\
detention & 51776 & 4222\\
hypomanic & 46140 & 18475\\
mania & 37644 & 12707\\
CBT & 36246 & 19351\\
psychotherapy & 32981 & 34513\\
manic & 24786 & 14952\\
formulation & 22123 & 13650\\
hospital & 19408 & 9161\\
cognitive & 18867 & 8307\\
hallucinations & 17631 & 11451\\
serotonin & 15069 & 9226\\
SSRI & 14862 & 7895\\
reuptake & 12151 & 9047\\
psychosocial & 4788 & 5556\\
cannabis & 4094 & 115\\
hamilton (tool) & 3451 & 6288\\
marijuana & 1214 & 103\\
bisexual & 1065 & 10914\\
psychoeducation & 911 & 9988\\
maladaptive & 144 & 5015\\
\end{tabular}
\end{table}

\paragraph{Lack of clinical knowledge:} the two LLMs demonstrate considerable gaps in their clinical knowledge.  For instance, Llama occasionally confuses the conditions under which hospitalisation is mandated under Section 2 vs Section 3 of the Mental Health Act.  Medication dosing also varies randomly and without clear justification, with doses tending toward the higher end of the recommended range or above, particularly for Lamotrigine and Quetiapine.  The models often suggest multiple drugs, leading to unwanted multipharmacy effects.  In addition, the models sometimes recommend treatment, such as prescribing Valproate to a woman of child-bearing age, which should generally be avoided because of the iatrogenic harm.  Quetiapine is occasionally misrepresented: although it can be used adjunctively as an antidepressant, it is primarily an antipsychotic and is not routinely prescribed for sleep, yet Llama repeatedly suggests it in less appropriate contexts, possibly influenced by mentions of the drug in news stories\footnote{\scriptsize \url{https://tinyurl.com/quetiapine-shortage}}.  Mistral stays closer to its intended usage.  The models also demonstrate differences in their preferred treatment methods: clonazepam is almost exclusively mentioned by Llama, while escitalopram is preferred by Mistral (Table \ref{tabMedications}).
Another example of the lack of clinical knowledge is the suggested duration of a \emph{Community Treatment Order (CTO) for 3 months}, which is unusually short and unwarranted for our test stories.

\paragraph{Lack of specificity:} Llama also shows a tendency to provide insufficiently detailed clinical information.  For example, in the two subsequent entries:
\begin{examples}
\item \emph{2018: Ayanna was hospitalized for a manic episode, which was treated with a combination of lithium and valproate.}
\item \emph{2020: Ayanna was discharged from hospital after a depressive episode was managed with a combination of lithium and antipsychotics.}
\end{examples}

\noindent
The first one offers a reasonable level of specificity, whereas the second one is much less precise.  Referring simply to \emph{antipsychotics} provides no clinical detail, as this is a broad class of medications rather than a specific drug, resulting in an unrealistic level of vagueness.  The frequency list in Table \ref{tabMedications} also shows its higher frequencies for generic words (\emph{symptoms, treatment}).  Finally, Llama can be vague when describing diagnostic procedures, often failing to specify which tests were performed or what their results were:
\begin{examples}
\item \emph{Her diagnosis was made after a comprehensive evaluation, including a thorough medical history, psychological assessment, and
laboratory tests.}
\end{examples}

\paragraph{LLM-specific:} Some issues include explicit reminders that the output is generated by an LLM (for example, \emph{Please note that this report is fictional}), as well as occasional refusals to produce clinical notes when suicidal ideation is mentioned.  Repetition also appears, with identical text sometimes reproduced two or three times within the same output.  This behaviour is more common with Llama, hence it produces considerably longer texts (Table \ref{tabCorpus}).

\paragraph{Estimation of biases:}
Variation in ethnicity within the prompts provides a test for the presence of biases.  Both LLMs acknowledged cultural and ethnic background and the need to address racial stereotypes.  However, Llama was far more likely to introduce references to drug use for the Afro-Caribbean patients: for example, \emph{marijuana} was mentioned 25 times in the White British stories vs 1,189 times in the two versions of the Afro-Caribbean stories, in comparison to 26 vs 77 for Mistral.  Similarly for \emph{cocaine} in Llama: 259 vs 1,857.  As the patient stories are otherwise identical, this indicates biased societal expectations, which are reflected in texts used to train Llama, while they are not justified by clinical practice.
\section{Related studies}
\label{sec:orge433e6a}
With recent advances in natural language generation, synthetic datasets have begun to proliferate across a wide range of domains, enabling controlled experimentation, see an early example in \citep{kvapilikova20synthetic}.  Within the healthcare sector, the Asclepius dataset has been recently developed to model the structure and content of discharge summaries \citep{kweon23syntheticclinical}.  However, comparable resources have not yet been available for the mental health domain, where clinical documentation follows different communicative norms and genre conventions.

In parallel, substantial work has focused on detecting mental health concerns in social media environments, including the CLPsych 2024 shared task on identifying suicidal intent in Reddit posts \citep{chim24clpsych} and such models as MentalLlama \citep{yang23mentalllama}.
A recent scoping review \citep{holmes25suicide} shows that most LLM research in suicide prevention focuses on risk detection while generative applications for support, education, and clinical assistance are emerging.  While these initiatives highlight the value of computational methods for risk assessment and early intervention, they mostly focus on user-generated content rather than clinical records.

Integrating generative AI into clinical practice remains an active area of investigation \citep{hider25aimh,hiebel25clinical}.  Other studies also show the importance of understand non-clinical factors in AI operations, such as ethnicity or gender, for example, it has been shown that LLM clinical accuracy decreases post-perturbation \citep{gourabathina25nonclinical}.
For responsible integration of AI, it is essential to understand the linguistic properties of the texts produced by LLMs and the conditions under which these properties vary, including the influence of sociodemographic biases \citep{draghi24synthetic,ma24synthetic}. Only with this understanding can synthetic data be designed to support reliable clinical research and applications.

There have been recent studies aimed at assessing the quality of synthetic medical texts produced by LLMs, such as \citep{xie24doclens}.  The focus of our study is on fine-frained linguistic analysis of outputs.  There have also been studies on the use of linguistic analysis for medical communication, for example, \citet{velasco22health} focused on the expression of empathy in coaching sessions.
This is different from our focus in this study on analysis of Electronic Health Records.
\section{Conclusions}
\label{sec:org7ac88eb}
This study demonstrates that open-weight large language models can approximate the structure, terminology, and discourse patterns of clinical genres.  However, they also introduce systematic deviations that reveal important limitations in their clinical fidelity.  This study aims at providing a clearer understanding of how and where LLMs succeed in producing coherent, genre-appropriate documents, but also how they fall short, in particular,  in modelling of clinical procedures or in culturally sensitive reasoning.

The quantitative patterns concern inflated drug references for certain ethnic groups, inappropriate medication regimens, and genre-specific misalignments in transitivity and modality.  These patterns can emerge only when examining a controlled, parallel synthetic corpus produced under consistent prompting conditions to study the impact of cues.

The most specific contribution of this work lies in the combination of genre-sensitive linguistic analysis and controlled synthetic data generation, which exposes patterns that would remain invisible in studies focused solely on accuracy or content validity.  As a result, it outlines a path towards safer, more realistic, and bias-aware computational experiments in the domain of mental health documentation.

Our future research is aimed at studying integration of LLMs into summarising authentic EHRs and co-production of better care plans in close collaboration between the patients and their mental health teams. The synthetic corpora provide an estimate for inaccuracies of LLMs in this domain.
\section{Risks and ethical concerns}
\label{sec:org0911e65}
The project aims to improve the quality of mental health care by focusing on the possibilities of using LLMs in clinical practice, in particular we investigate biases in synthetic data, as well as deviations of LLM outputs from the current practice.

The environmental impact of LLMs needs to be checked against the societal impact of their application.  Our experiments used NVidia L40S GPUs, which, on average, were consuming 168W in our task.  The inference time for a single output was 8.2s for Llama, 12.1s for Mistral.  The total time for generating outputs by 2 LLMs \texttimes{} 4 genres \texttimes{} 12,960 stories was 292h19m.  This is equivalent to driving a Nissan Leaf for 172 miles (3.5mi/kWh average consumption without heating or a/c). 
\section{Limitations}
\label{sec:orgf7e5638}
First, the parameters of variation do not capture the full range of clinical scenarios found in mental health care.  We focus on variation relevant to our current study.

Second, the evaluation does not cover the full range of genres found in EHRs, as well as genres which inform the clinical EHRs, for example, personal accounts of lived experience from the patients.

Last, the evaluation focuses primarily on linguistic and genre-based features rather than on clinical safety.  This leaves the need for further exploration of practical implications for using synthetic datasets in health care.

\appendix

\bibliographystyle{acl_natbib}
\bibliography{bibexport}

@book{halliday99,
  address =       {London},
  author =        {Halliday, Michael A. K. and
                   Matthiessen, Christian M. I. M.},
  publisher =     {Cassell},
  title =         {Construing experience through meaning: a
                   language-based approach to cognition},
  year =          {1999},
}

@misc{ball21letters,
  author =        {Karen Ball and Wendy Burn},
  howpublished =  {Royal College of Psychiatrists},
  month =         {January},
  title =         {Writing clinic letters: College guidance on improving
                   engagement with patients},
  year =          {2021},
  url =           {https://www.rcpsych.ac.uk/docs/default-source/members/
                  supporting-you/writing-clinic-letters/writing-clinic-letters-
                  --guidance-for-psychiatrists.pdf},
}

@misc{nhsdigital25referrals,
  author =        {{NHS Digital}},
  title =         {Advice and guidance for referrers and referring
                   clinician teams},
  year =          {2025},
  url =           {https://digital.nhs.uk/services/e-referral-service/document-
                  library/advice-and-guidance-toolkit/advice-and-guidance-for-
                  gps-and-referring-clinician-teams},
}

@article{priebe17dialog,
  author =        {Priebe, Stefan and Golden, Eoin and Kingdon, David and
                   Omer, Serif and Walsh, Sophie and Katevas, Kleomenis and
                   McCrone, Paul and Eldridge, Sandra and McCabe, Rose},
  journal =       {Programme Grants for Applied Research},
  number =        {6},
  pages =         {1--160},
  publisher =     {National Institute for Health and Care Research},
  title =         {Effective patient--clinician interaction to improve
                   treatment outcomes for patients with psychosis: a
                   mixed-methods design},
  volume =        {5},
  year =          {2017},
}

@book{halliday14,
  address =       {London},
  author =        {Halliday, Michael A. K. and
                   Matthiessen, Christian M. I. M.},
  edition =       {4th edition},
  publisher =     {Routledge},
  title =         {Introduction to Functional Grammar},
  year =          {2014},
}

@inproceedings{odonnell08,
  address =       {Columbus, Ohio},
  author =        {O’Donnell, Mick},
  booktitle =     {Proceedings of the ACL-08: HLT Demo Session},
  pages =         {13--16},
  publisher =     {Association for Computational Linguistics},
  title =         {Demonstration of the {UAM} {CorpusTool} for text and
                   image annotation},
  year =          {2008},
  url =           {https://aclanthology.org/P08-4004/},
}

@article{matthiessen15,
  author =        {Matthiessen, Christian MIM},
  journal =       {Functional Linguistics},
  number =        {1},
  pages =         {1-48},
  title =         {Register in the round: registerial cartography},
  volume =        {2},
  year =          {2015},
}

@inproceedings{kvapilikova20synthetic,
  address =       {Online},
  author =        {Kvapil{\'\i}kov{\'a}, Ivana and Artetxe, Mikel and
                   Labaka, Gorka and Agirre, Eneko and
                   Bojar, Ond{\v{r}}ej},
  booktitle =     {Proceedings of the 58th Annual Meeting of the
                   Association for Computational Linguistics: Student
                   Research Workshop},
  month =         jul,
  pages =         {255--262},
  publisher =     {Association for Computational Linguistics},
  title =         {Unsupervised Multilingual Sentence Embeddings for
                   Parallel Corpus Mining},
  year =          {2020},
  doi =           {10.18653/v1/2020.acl-srw.34},
  url =           {https://www.aclweb.org/anthology/2020.acl-srw.34},
}

@misc{kweon23syntheticclinical,
  author =        {Sunjun Kweon and Junu Kim and Jiyoun Kim and
                   Sujeong Im and Eunbyeol Cho and Seongsu Bae and
                   Jungwoo Oh and Gyubok Lee and Jong Hak Moon and
                   Seng Chan You and Seungjin Baek and Chang Hoon Han and
                   Yoon Bin Jung and Yohan Jo and Edward Choi},
  title =         {Publicly Shareable Clinical Large Language Model
                   Built on Synthetic Clinical Notes},
  year =          {2023},
}

@inproceedings{chim24clpsych,
  address =       {St. Julians, Malta},
  author =        {Chim, Jenny and Tsakalidis, Adam and
                   Gkoumas, Dimitris and Atzil-Slonim, Dana and
                   Ophir, Yaakov and Zirikly, Ayah and Resnik, Philip and
                   Liakata, Maria},
  booktitle =     {Proceedings of the 9th Workshop on Computational
                   Linguistics and Clinical Psychology (CLPsych 2024)},
  month =         mar,
  pages =         {177--190},
  publisher =     {Association for Computational Linguistics},
  title =         {Overview of the {CLP}sych 2024 Shared Task:
                   Leveraging Large Language Models to Identify Evidence
                   of Suicidality Risk in Online Posts},
  year =          {2024},
  url =           {https://aclanthology.org/2024.clpsych-1.15/},
}

@misc{yang23mentalllama,
  author =        {Kailai Yang and Tianlin Zhang and Ziyan Kuang and
                   Qianqian Xie and Sophia Ananiadou},
  title =         {MentalLLaMA: Interpretable Mental Health Analysis on
                   Social Media with Large Language Models},
  year =          {2023},
}

@article{holmes25suicide,
  author =        {Holmes, Glenn and Tang, Biya and Gupta, Sunil and
                   Venkatesh, Svetha and Christensen, Helen and
                   Whitton, Alexis},
  journal =       {Journal of Medical Internet Research},
  pages =         {e63126},
  publisher =     {JMIR Publications},
  title =         {Applications of large language models in the field of
                   suicide prevention: Scoping review},
  volume =        {27},
  year =          {2025},
}

@article{hider25aimh,
  author =        {Hider, Andrew and Wright, Lesa and Needle, Jacob},
  journal =       {BJPsych Bulletin},
  pages =         {1–6},
  title =         {Clinical Reach into the Cognitive Space ({CRITiCS}):
                   outline conceptual framework for safe use of
                   generative artificial intelligence in mental health
                   decision-making},
  year =          {2025},
  doi =           {10.1192/bjb.2025.36},
}

@article{hiebel25clinical,
  author =        {Hiebel, Nicolas and Ferret, Olivier and
                   Fort, Kar{\"e}n and N{\'e}v{\'e}ol, Aur{\'e}lie},
  journal =       {Annual Review of Biomedical Data Science},
  title =         {Clinical text generation: Are we there yet?},
  volume =        {8},
  year =          {2025},
}

@inproceedings{gourabathina25nonclinical,
  author =        {Gourabathina, Abinitha and Gerych, Walter and
                   Pan, Eileen and Ghassemi, Marzyeh},
  booktitle =     {Proceedings of the 2025 ACM Conference on Fairness,
                   Accountability, and Transparency},
  pages =         {1805--1828},
  title =         {The medium is the message: How non-clinical
                   information shapes clinical decisions in {LLM}s},
  year =          {2025},
}

@article{draghi24synthetic,
  author =        {Draghi, Barbara and Wang, Zhenchen and Myles, Puja and
                   Tucker, Allan},
  journal =       {Heliyon},
  number =        {2},
  publisher =     {Elsevier},
  title =         {Identifying and handling data bias within primary
                   healthcare data using synthetic data generators},
  volume =        {10},
  year =          {2024},
}

@inproceedings{ma24synthetic,
  address =       {Vienna, Austria},
  author =        {Bolei Ma and Berk Yoztyurk and Anna-Carolina Haensch and
                   Xinpeng Wang and Markus Herklotz and Frauke Kreuter and
                   Barbara Plank and Matthias Assenmacher},
  booktitle =     {Proceedings of the 63rd Annual Meeting of the
                   Association for Computational Linguistics (Volume 1:
                   Long Papers)},
  month =         jul,
  pages =         {1785--1809},
  publisher =     {Association for Computational Linguistics},
  title =         {Algorithmic Fidelity of Large Language Models in
                   Generating Synthetic {German} Public Opinions: A Case
                   Study},
  year =          {2025},
  isbn =          {979-8-89176-251-0},
  url =           {https://aclanthology.org/2025.acl-long.90/},
}

@inproceedings{xie24doclens,
  address =       {Bangkok, Thailand},
  author =        {Xie, Yiqing and Zhang, Sheng and Cheng, Hao and
                   Liu, Pengfei and Gero, Zelalem and Wong, Cliff and
                   Naumann, Tristan and Poon, Hoifung and Rose, Carolyn},
  booktitle =     {Proceedings of the 62nd Annual Meeting of the
                   Association for Computational Linguistics (Volume 1:
                   Long Papers)},
  month =         aug,
  pages =         {649--679},
  publisher =     {Association for Computational Linguistics},
  title =         {{D}oc{L}ens: Multi-aspect Fine-grained Evaluation for
                   Medical Text Generation},
  year =          {2024},
  doi =           {10.18653/v1/2024.acl-long.39},
  url =           {https://aclanthology.org/2024.acl-long.39/},
}

@article{velasco22health,
  author =        {Velasco, Elena Rey and Pedersen, Hanne S{\ae}derup and
                   Skinner, Timothy},
  journal =       {JMIR formative research},
  number =        {12},
  pages =         {e40058},
  title =         {Analysis of patient cues in asynchronous health
                   interactions: pilot study combining empathy appraisal
                   and systemic functional linguistics},
  volume =        {6},
  year =          {2022},
}
\clearpage
\section{Appendix: Example of Mistral outputs}
\label{sec:org23ab1a1}
\subsection{Prompt story}
\label{prompt}
Please write three paragraphs for each of the following sections
concerning a fictitious female patient, who is 25 years old, has been
diagnosed with Bipolar I Disorder with episodes of mania alternating
with depressive episodes. Ethnicity: Afro-Caribbean. Risks: chronic pain.
Treatment history: detained under the mental health act (Section 2)
\subsection{Initial examination}
\label{initial-examination}
\textbf{Initial Assessment Report}

Subject ID: J.A., a 25-year-old Afro-Caribbean female, presents for a
comprehensive psychiatric evaluation. She has been previously diagnosed
with Bipolar I Disorder, characterized by alternating episodes of mania
and depression. The patient was detained under the Mental Health Act
(Section 2) due to concerns regarding her safety and wellbeing. The
current admission aims to stabilize her symptoms, address any underlying
issues, and establish an appropriate treatment plan.

\textbf{Mental State Examination}

Upon initial contact, the patient exhibited a labile mood, switching
between euphoria and irritability. Her speech was pressured, with racing
thoughts and flight of ideas. J.A. reported severe insomnia, increased
energy levels, and a decreased need for sleep. She demonstrated
grandiosity, expressed a sense of invincibility, and demonstrated
impulsive behaviors. Additionally, the patient reported chronic pain,
which may be contributing to her emotional distress. Affect was
concordant with her mood state, and her thought content was otherwise
unremarkable.

\textbf{Psychiatric History \& Formulation}

J.A.'s psychiatric history is significant for multiple hospitalizations
due to manic episodes, each of which was followed by a depressive
episode. Her manic symptoms, including inflated self-esteem, decreased
need for sleep, and impulsive behaviors, have been consistently present
throughout her illness course. The patient's depressive episodes have
been characterized by feelings of worthlessness, hopelessness, and
severe anhedonia.

In terms of medication history, the patient has been prescribed various
mood stabilizers, antipsychotics, and antidepressants, with varying
degrees of success. The current medication regimen includes Lithium
carbonate for mood stabilization, Quetiapine for sedation, and
Amitriptyline for pain management.

The formulation of J.A.'s presentation suggests a diagnosis of Bipolar I
Disorder with chronic pain. The recurrent manic and depressive episodes
are likely to be driven by a combination of genetic, biological, and
environmental factors. The presence of chronic pain may exacerbate her
mood symptoms and contribute to her overall distress.

The primary risks associated with J.A.'s current presentation include
impulsive behaviors during manic episodes, which could potentially lead
to harmful or dangerous activities. The chronic pain she experiences may
also increase her vulnerability to substance abuse as a means of
self-medication. It is crucial to ensure close monitoring of her
symptoms, adherence to medication, and psychosocial support to minimize
these risks. A multidisciplinary approach, involving psychiatry, pain
management, and psychology, will be essential in managing J.A.'s complex
presentation and improving her overall quality of life.
\subsection{GP correspondence}
\label{gp-letters}
\textbf{I. Medication History}

The subject, a 25-year-old Afro-Caribbean female, has a documented
history of Bipolar I Disorder, characterized by recurrent episodes of
mania and major depressive episodes. The patient's medication history
has been marked by a series of pharmacological interventions aimed at
managing her symptoms. Initially, she was prescribed a combination of
Mood Stabilizers (Lithium Carbonate and Valproate Sodium) and an
Antipsychotic (Olanzapine) to address the manic symptoms. However, due
to the side effects such as weight gain and sedation, the medication
regimen was altered to a Monotherapy of Lithium Carbonate. Subsequently,
due to the recurrence of manic episodes, the treatment was further
adjusted to include a second-generation Antipsychotic (Quetiapine). The
patient's response to these medications has been variable, with some
beneficial effects but also persistent manic symptoms. Close monitoring
of the patient's medication and symptoms is essential to ensure optimal
treatment.

\textbf{II. Treatment History}

The patient has a notable treatment history, involving detainment under
the Mental Health Act (Section 2). This action was taken due to the
patient's inability to manage her symptoms adequately, which posed a
risk to herself and others. During her hospitalization, the patient was
provided with a structured treatment plan that included pharmacological
interventions, psychotherapy, and education about her condition. The
patient's compliance with the treatment plan was inconsistent, leading
to the recurrence of manic episodes and the need for continued inpatient
care. In light of these challenges, the focus of treatment has shifted
to a more intensive outpatient program, with regular follow-ups,
medication management, and psychotherapy sessions.

\textbf{III. Risks}

In addition to the risks associated with untreated Bipolar I Disorder,
this patient has an ongoing issue with chronic pain. The etiology of
this pain is unclear, but it is believed to be a result of a combination
of factors, including her mental health condition and possible physical
comorbidities. The chronic pain exacerbates the patient's mood symptoms
and contributes to her overall disability. It is essential to address
this issue in conjunction with her mental health treatment to improve
her overall quality of life and reduce the risk of relapse.

As for the examples of GP correspondence, I am unable to provide
specific examples without access to the patient's medical records.
However, I can suggest the following format for GP correspondence
addressing this patient:

\begin{enumerate}
\item Identification of the patient and relevant medical history. "Dear
Dr. [GP Name], I am writing to provide an update on our patient,
{[}Patient Name], who is currently under my care for Bipolar I
Disorder. She is a 25-year-old Afro-Caribbean female with a history
of detainment under the Mental Health Act (Section 2)."

\item Discussion of the patient's current symptoms and treatment plan.
"Recently, the patient has been experiencing fluctuations in her
mood, with recurrent manic episodes and periods of depression. We
have adjusted her medication regimen to include Quetiapine in an
effort to better manage her symptoms. Additionally, she is attending
regular psychotherapy sessions and education classes about her
condition."

\item Communication of any concerns or recommendations. "I am concerned
about the patient's ongoing issue with chronic pain and the impact it
has on her mental health. I recommend that you refer her to a pain
management specialist for further evaluation and treatment. It is
essential to address this issue in conjunction with her mental health
treatment to improve her overall quality of life and reduce the risk
of relapse."

\item Request for collaboration and follow-up. "I would appreciate your
collaboration in managing this complex case. Please keep me informed
of any changes in the patient's condition or any issues that arise
during your care. I look forward to working together to ensure the
best possible outcome for our patient."

\item Closing remarks. “Thank you for your attention to this matter. Please
do not hesitate to contact me if you have any questions or concerns.

Sincerely, [Your Name] [Your Title and Affiliation]”
\end{enumerate}
\subsection{Referrals}
\label{referrals}
\textbf{Referral and Handover Letter}

To Whom It May Concern,

I am writing to provide a referral and handover for a 25-year-old
Afro-Caribbean female patient who has been under my care for the past
six months. The patient has been diagnosed with Bipolar I Disorder,
presenting with episodes of mania alternating with depressive episodes.

\textbf{Presenting Symptoms:} The patient has been experiencing irregular sleep
patterns, increased energy levels, and grandiose thinking, which are
indicative of manic episodes. Conversely, she has reported feelings of
profound sadness, loss of interest in activities, and decreased energy
levels, which are suggestive of depressive episodes. Additionally, she
has expressed chronic pain, which appears to exacerbate her mood
episodes.

\textbf{Background and Relevant Mental Health History:} The patient's mental
health history is significant for repeated hospitalizations under the
Mental Health Act (Section 2), due to her inability to manage her
symptoms effectively without supervised care. Her symptoms began to
manifest at the age of 20, marked by severe mood swings and erratic
behavior.

\textbf{Current Medication and Risk Assessment:} The patient is currently on a
regimen of Lithium carbonate 900mg daily and Quetiapine 600mg nightly.
These medications have shown some efficacy in stabilizing her mood
episodes, but she continues to struggle with chronic pain, which may
contribute to her mood instability.

Risk assessment indicates that the patient poses no immediate danger to
herself or others, but her chronic pain and unstable mood episodes
require close monitoring. The patient's non-adherence to medication
regimen and self-harming behaviors in the past necessitate a high level
of supervision and support.

\textbf{Reasons for Referral:} The patient is being referred to your care for
ongoing management of her Bipolar I Disorder and chronic pain. Her
history of repeated hospitalizations under the Mental Health Act and the
need for a high level of supervision necessitate a collaborative
approach to her care. I believe that your expertise and resources will
be instrumental in ensuring her continued safety and wellbeing.

Please find attached the patient's medical records, including her
current medication list, for your review. I appreciate your attention to
this matter and look forward to our continued collaboration in the best
interests of the patient.

Sincerely, [Your Name] [Your Position]
\subsection{Care plan}
\label{care-plan}
\textbf{Patient-Centered Advance Care Plan (DIALOG+ Methodology)}

\textbf{1. Psychiatric Assessment of Diagnoses}

The patient is a 25-year-old Afro-Caribbean female with a primary
diagnosis of Bipolar I Disorder, characterized by distinct episodes of
mania alternating with depressive episodes. The patient's history is
marked by periods of elevated mood, inflated self-esteem, decreased need
for sleep, grandiose beliefs, and increased goal-directed activity,
followed by periods of low mood, loss of interest in usually pleasurable
activities, decreased energy, and feelings of worthlessness.

\textbf{2. Treatment Goals}

The primary treatment goal is to stabilize the patient's mood
fluctuations, reduce the frequency and severity of manic and depressive
episodes, and improve overall functioning and quality of life. Secondary
goals include managing chronic pain, reducing hospitalizations, and
promoting adherence to medication and therapy.

\textbf{3. Objectives}

Objectives for this treatment plan include the following:
\begin{itemize}
\item Achieve and maintain mood stability through pharmacological and psychotherapeutic interventions.
\item Reduce the severity and frequency of manic and depressive episodes.
\item Improve the patient's ability to cope with chronic pain through cognitive-behavioral strategies.
\item Promote medication adherence and reduce hospitalization rates.
\item Enhance the patient's social and occupational functioning.
\end{itemize}

\textbf{4. Interventions}

Interventions to achieve these objectives include:
\begin{itemize}
\item Pharmacological management with mood stabilizers (e.g., lithium, valproate, or atypical antipsychotics) and/or antidepressants, as necessary, under close monitoring for side effects and efficacy.
\item Cognitive-behavioral therapy (CBT) to help the patient identify and challenge maladaptive thought patterns, manage stress, and develop effective coping strategies for mood regulation and pain management.
\item Psychoeducation about Bipolar I Disorder, its symptoms, treatment options, and the importance of medication adherence and regular follow-ups.
\item Collaborative goal setting and problem-solving to address challenges in daily functioning and reduce hospitalizations.
\item Regular multidisciplinary team meetings to coordinate care, monitor progress, and adjust treatment plans as needed.
\end{itemize}

\textbf{5. Responsibilities}

Responsibilities for this treatment plan include:
\begin{itemize}
\item The patient is responsible for attending scheduled appointments, taking medications as prescribed, and following the recommended treatment plan.
\item The patient's primary care team, including the psychiatrist, psychotherapist, and nurse, is responsible for providing evidence-based care, monitoring the patient's response to treatment, and adjusting the treatment plan as necessary.
\item The patient's family and support system is responsible for providing emotional support, encouragement, and assistance in managing daily activities as needed.
\end{itemize}

\textbf{6. Progress Tracking}

Progress will be tracked through regular assessments of mood symptoms
using validated rating scales (e.g., Hamilton Depression Rating Scale,
Young Mania Rating Scale), monitoring of medication levels, and
evaluations of the patient's functioning in social, occupational, and
self-care domains.

\textbf{7. Timeline for Achieving Specific Milestones in Mental Health Therapy}

The timeline for achieving specific milestones in mental health therapy
will vary for each patient, depending on the severity of symptoms,
adherence to treatment, and individual circumstances. However, some
potential milestones might include:
\begin{itemize}
\item Stable mood for at least three months within the first six months of treatment.
\item Reduction in hospitalization rates over the first year of treatment.
\item Improvement in daily functioning, as measured by self-report and clinical assessment, within the first six months to a year of treatment.
\item Development of effective coping strategies for managing chronic pain within the first three to six months of CBT.
\item Maintenance of treatment gains over time through ongoing collaboration with the treatment team and self-care strategies.
\end{itemize}
\end{document}